%% file: main.tex
\documentclass{IEEEcsmag-arxiv}

\usepackage[colorlinks,urlcolor=blue,linkcolor=blue,citecolor=blue]{hyperref}
\expandafter\def\expandafter\UrlBreaks\expandafter{\UrlBreaks\do\/\do\*\do\-\do\~\do\'\do\"\do\-}
\usepackage{upmath,color}
\usepackage[export]{adjustbox}

\usepackage{graphicx}
\usepackage{subcaption}
\usepackage{amsfonts}

\pubyear{2026}

\newcommand{\fdist}{$f(d)$} 

\begin{document}

\sptitle{Article Type: Feature}

\title{Pre-Deployment Complexity Estimation for Federated Perception Systems}

\author{KMA Solaiman}
\affil{ksolaima@umbc.edu, University of Maryland Baltimore County, Baltimore, MD 21250, USA} 

\author{Shafkat Islam}
\affil{islam59@pnw.edu, Purdue University Northwest, Hammond, IN 46323, USA}

\author{Ruy de Oliveira}
\affil{rdeolive@purdue.edu, Purdue University, West Lafayette, IN 47907, USA}

\author{Bharat Bhargava}
\affil{bbshail@purdue.edu, Purdue University, West Lafayette, IN 47907, USA}


\begin{abstract}
Edge Artificial Intelligence (AI) systems increasingly rely on federated learning to train perception models in distributed, privacy-preserving, and resource-constrained environments. Before training, however, practitioners often lack practical tools for estimating task difficulty in terms of expected accuracy and communication effort. We present a classifier-agnostic, pre-deployment framework that combines intrinsic data properties such as dimensionality, sparsity, and heterogeneity, with client-distribution composition to estimate learning complexity in federated perception systems. Using federated learning as a representative distributed training setting, we examine how learning difficulty varies across different federated configurations. Experiments on three MNIST variants show strong negative correlations between the combined complexity metric and maximum and average federated accuracy, while the intrinsic and distributed components exhibit consistent relationships with communication effort. These findings suggest that complexity estimation can serve as a practical diagnostic tool for resource planning, dataset assessment, and feasibility evaluation in edge-deployed perception systems.

\end{abstract}

\maketitle

\section{INTRODUCTION}

Deploying perception models in edge environments introduces challenges that extend beyond model accuracy alone. Practical deployments must account for communication limitations, constrained computation, and fragmented data ownership, particularly when training data cannot be centralized. Distributed training paradigms have therefore become central to modern Edge AI systems, motivating the need for tools that help practitioners reason about training feasibility and cost before large-scale learning is attempted.


Federated learning (FL) has emerged as a widely adopted framework for distributed and privacy-preserving model training, especially in perception applications \cite{FL2020}. While a substantial body of research has focused on improving federated optimization strategies and convergence behavior \cite{44822, reddi2020adaptive}, considerably less attention has been paid to understanding why certain federated learning tasks are inherently more difficult than others. In practice, federated tasks trained using identical models and algorithms can exhibit markedly different convergence rates, communication costs, and final accuracy, differences that are often discovered only after training has already begun.

A key factor underlying this variability is the intrinsic structure of the data and the characteristics of the distributed environment. Prior work has shown that properties such as dimensionality, sparsity, and heterogeneity play a critical role in determining learning difficulty in perception domains \cite{Scheidegger2021, Pereyda2020MeasuringDC, Krusinga2019}. However, most existing approaches assess complexity from the perspective of a single dataset or a centralized learning agent. Extending such analysis to distributed environments remains challenging, as complexity in these settings depends not only on local data properties but also on how data are partitioned across multiple entities.\\

In this work, we adopt a measurement-oriented perspective and address the problem of \emph{pre-deployment complexity estimation} for federated perception systems. Rather than proposing a new learning algorithm or federated optimization policy, our objective is to quantify how difficult a federated learning task is likely to be based solely on intrinsic data properties and distributed environment characteristics. This diagnostic view complements existing work on federated optimization by providing insight into task difficulty prior to training.

We propose a classifier-agnostic complexity estimation framework that captures three intrinsic attributes of perception data — dimensionality, sparsity, and heterogeneity, and integrates them with properties of the federated environment, such as the number and composition of participating clients. Federated learning is used as a representative distributed training paradigm, enabling the analysis of different federated configurations without modifying the underlying learning process. The resulting complexity metric allows different federated learning configurations to be compared based on expected difficulty rather than observed performance.

To study the relationship between the proposed complexity metric and federated learning behavior, we conduct controlled experiments on multiple variants of the MNIST dataset in distributed settings. These experiments are designed to isolate complexity effects and examine how intrinsic and distributed complexity relate to federated accuracy and communication effort. 

\noindent The main contributions of this work are as follows:

\begin{itemize}
    \item We introduce a classifier-agnostic framework for estimating intrinsic complexity in perception datasets, based on dimensionality, sparsity, and heterogeneity, without reliance on classifier performance.
    \item We propose a unified complexity metric for federated learning environments that integrates intrinsic data properties with distributed environment characteristics, enabling comparison of federated training configurations.
    \item We empirically evaluate the framework on multiple MNIST dataset variants, demonstrating strong negative correlations between the combined metric and federated accuracy and consistent associations between its components and communication effort.
\end{itemize}



\section{BACKGROUND AND RELATED WORK} 

\subsection{\textbf{Complexity Estimation in Perception Domains}}
Estimating the complexity of perception datasets and learning tasks has been studied from several perspectives, including information theory, uncertainty estimation \cite{Batty2014, Langley2020, Cardoso2021}, and dataset organization. Prior work has shown that intrinsic properties such as dimensionality, sparsity, and heterogeneity play an important role in determining learning difficulty in perception tasks.

Pereyda et al.~\cite{Pereyda2020MeasuringDC} proposed a theoretical framework for measuring domain complexity and evaluated it using approximations derived from neural network models. Krusinga et al.~\cite{Krusinga2019} investigated dataset complexity by estimating probability densities of image distributions using generative adversarial networks, enabling the detection of outliers and domain shifts at the cost of significant computational overhead. Scheidegger et al.~\cite{Scheidegger2021} examined classification difficulty using clustering-based metrics and lightweight neural probes, demonstrating that simple models can approximate dataset difficulty efficiently.

While these approaches provide valuable insight into dataset complexity, they are typically developed for centralized settings and often rely on classifier behavior or learned representations. In contrast, our work focuses on estimating intrinsic complexity directly from data properties and extends the analysis to distributed environments 
by incorporating the composition of client label-support types.




\subsection{\textbf{Distributed and Federated Learning}}

Federated learning has emerged as a widely adopted framework for distributed and privacy-preserving model training \cite{44822}. A substantial body of research has focused on improving federated optimization and convergence behavior through methods such as FedAvg \cite{44822}, FedAdaGrad \cite{reddi2020adaptive} and FedYogi \cite{reddi2020adaptive}. These approaches aim to improve training efficiency and robustness under heterogeneous data distributions and system constraints.

However, existing federated learning methods primarily address how to train a model efficiently given a fixed task and dataset distribution. 
These methods generally do not provide an explicit pre-training estimate of federated task difficulty or model how intrinsic data complexity and client-distribution composition jointly characterize learning difficulty.
As a result, task feasibility and communication effort are typically assessed only after training is underway.

\subsection{\textbf{Positioning of This Work}}
This work complements existing research by introducing a classifier-agnostic, measurement-oriented framework for estimating learning complexity in federated perception systems. By combining intrinsic dataset-specific properties with distributed client label-support composition, the proposed approach provides a pre-deployment diagnostic of expected task difficulty. Federated learning is used as a representative distributed training setting; the metric neither modifies nor optimizes the learning algorithm, but instead supports feasibility assessment and resource-aware planning before training.
%

\section{COMPLEXITY ESTIMATION FRAMEWORK} 
\label{sec:framework}



This section introduces a \emph{pre-training, classifier-agnostic framework} for estimating the expected difficulty of federated perception tasks from intrinsic data properties and client-distribution composition. The framework separates complexity into two complementary components: intrinsic complexity, which characterizes properties of the client-local data, and distributed-environment complexity, which characterizes the fragmentation of clients among distinct data-distribution types.


Intrinsic data complexity is characterized using three complementary properties: dimensionality, sparsity-derived complexity, and heterogeneity. Dimensionality captures the observable feature-space size and latent structure of the data. Sparsity characterizes the
concentration of the data representation, with its complement used
as a complexity-aligned measure of representation density.
Heterogeneity captures within-image visual diversity in the observed data. These
properties are computed directly from the available data before
model training. 

Distributed-environment complexity is modeled using the federated learning paradigm. Federated learning provides a concrete setting in which data are distributed across multiple participating entities, and learning proceeds through iterative communication rounds. 
In this work, the distributed-environment complexity metric is determined from the number and frequencies of distinct client-distribution types. 
The two components are subsequently combined to obtain a federated learning complexity score.
%



\begin{figure*}[htbp]
\centering
\includegraphics[width=\linewidth, height=0.4\linewidth]{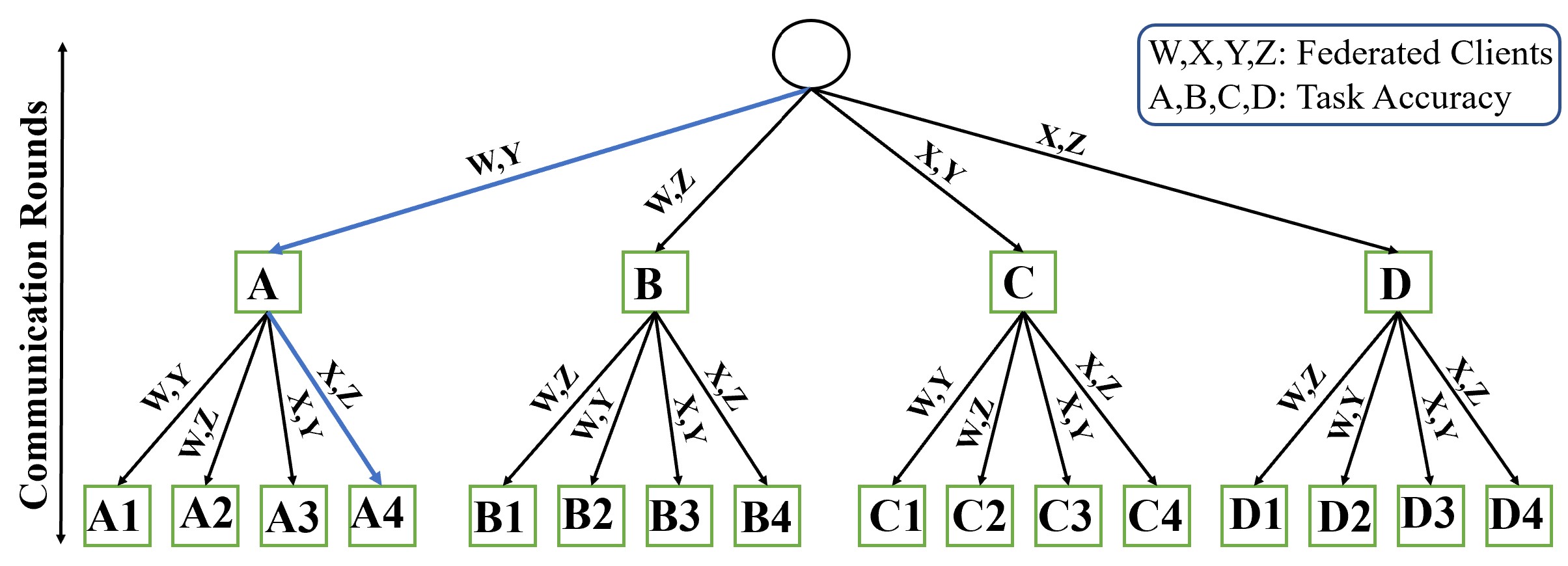}
\caption{Conceptual illustration of federated learning configurations
across communication rounds. Edge labels indicate participating
client subsets, nodes represent the resulting task-accuracy states,
and the highlighted route shows one possible participation path.
$W,X,Y,Z$ identify participating clients, and $A$ -- $D$ and $A1$ -- $D4$ index the
resulting task-accuracy states. 
}
\label{fig:fed_game_tree}
\end{figure*}

Figure~\ref{fig:fed_game_tree} provides a conceptual illustration
of federated configurations. Each path represents a possible sequence
of participating clients across communication rounds and illustrates
how differences in data exposure and aggregation order may lead to different learning
outcomes. 
Although the figure depicts possible client-participation paths, the proposed metric does not model path order or round-level participation.
Instead, it
characterizes a configuration using the intrinsic properties of the
client-local 
data and the composition of client label-support types.

This formulation enables different federated training configurations to be compared in terms of expected difficulty, accuracy trends, and communication effort, without requiring model training to be performed in advance.

\subsection{\textbf{Perception Domain Intrinsic Complexity}}
\label{sec:intrinsic-domain-complexity}

Intrinsic complexity refers to properties of the data distribution itself that influence learning difficulty independent of the learning model.
In perception tasks, these properties arise from the structure, variability, and data organization and can be assessed prior to training. In this work, intrinsic complexity is characterized along three complementary axes: dimensionality, sparsity, and heterogeneity.

\vspace{0.5em}
\textit{\textbf{Dimensionality.}}
Let $\mathcal{D}$ denote a dataset containing $N_s$ observations,
and let $X\in\mathbb{R}^{N_s\times N_f}$ denote its feature matrix, where each observation is represented by $N_f$ ambient features.
To characterize how many coordinates of this
representation exhibit non-negligible variation across the dataset, we
use a variance-based feature filter.
For feature \(q\), its empirical variance is defined as $\widehat{\sigma}_q^2$.
Given a non-negative threshold \(v_\theta\), the
variance-filtered ambient feature count is
\begin{equation}
    N_f^r(v_\theta)
    =
    \sum_{q=1}^{N_f}
    \mathbb{I}
    \left[
        \widehat{\sigma}_q^2>v_\theta
    \right],
\end{equation}
where \(\mathbb{I}[\cdot]\) is the indicator function. 
However, the resulting count remains dependent on the original feature
representation, pixel scaling, and the selected threshold.
Consequently, \(N_f^r(v_\theta)\) is interpreted as a complementary 
representation-dependent ambient-space diagnostic rather than as a
measure of intrinsic dimensionality.

In contrast, perception data commonly occupy a lower-dimensional manifold embedded within this ambient feature space. 
Since intrinsic dimensionality (ID) estimates the effective
degrees of freedom associated with the local geometry of the data, we treat ID as the primary measure of effective dimensionality. 

Let $\{\mathbf{x}_i\}_{i=1}^{N_s}$ denote the rows of $X$, and let
$T_j(\mathbf{x})$ denote the distance from observation
$\mathbf{x}$ to its $j$-th nearest neighbor. 
Following the maximum-likelihood estimator of Levina and Bickel~\cite{levina2004maximum}, the local intrinsic dimensionality at \(x\) is estimated as
\begin{equation}
    \widehat{id}_{\kappa}(\mathbf{x})
    =
    \left[
        \frac{1}{{\kappa}-1}
        \sum_{j=1}^{{\kappa}-1}
        \log
        \left(
            \frac{T_{\kappa}(\mathbf{x})}
                 {T_j(\mathbf{x})}
        \right)
    \right]^{-1}.
\end{equation}
A dataset-level intrinsic-dimensionality estimate is obtained by aggregating the local estimates across the evaluated samples:
\(
    \widehat{\operatorname{ID}}(\mathcal{D})
    =
    \frac{1}{N_s}
    \sum_{i=1}^{N_s}
    \widehat{id}_{\kappa}(\mathbf{x}_i).
\)





\vspace{0.5em}
\textit{\textbf{Sparsity.}}
Sparsity characterizes the extent to which the intensity of an image
is concentrated in a limited number of pixel coordinates. For image
\(I_i\), let
\(
\mathbf{x}_i\in\mathbb{R}_{\geq0}^{N_f}
\)
denote its flattened nonnegative pixel-intensity vector. We measure
image-level sparsity using
\begin{equation}
    s(\mathbf{x}_i)
    =
    \frac{
        \sqrt{N_f}
        -
        \lVert\mathbf{x}_i\rVert_1/
        \lVert\mathbf{x}_i\rVert_2
    }{
        \sqrt{N_f}-1
    }.
\end{equation}
The measure is bounded by \(0\leq s(\mathbf{x}_i)\leq1\). A value near
one indicates that image intensity is concentrated in relatively few
pixels, whereas a value near zero indicates that intensity is
distributed more uniformly across the ambient representation.
%
We define the dataset-level sparsity metric as the mean image-level score:
\(
    S(\mathcal{D})
    =
    \frac1{N_s}
    \sum_{i=1}^{N_s}
    s(\mathbf{x}_i).
\)


To align sparsity with the direction of the other complexity
indicators, we use the complementary quantity as sparsity-derived complexity:
$C_S(\mathcal D)=1-S(\mathcal D)$, for which larger values indicate
denser pixel activation and greater representation density.




\vspace{0.5em}
\textit{\textbf{Heterogeneity.}}
Heterogeneity characterizes the diversity of pixel-intensity distributions observed within a perception dataset, which can influence learning difficulty.
We quantify heterogeneity using Shannon entropy \cite{shannon1948mathematical}, a standard information-theoretic measure of uncertainty.
For a grayscale image \(I_i\), let \(p_{ig}\) denote the proportion of its pixels having intensity \(g\in\{0,\ldots,255\}\).
We calculate the image-level Shannon
entropy as
\begin{equation}
    H(I_i)
    =
    -\sum_{g=0}^{255}
    p_{ig}\log_2p_{ig},
\end{equation}
where terms for which \(p_{ig}=0\) are omitted.
%
The dataset-level heterogeneity metric
is defined as the mean image-level entropy:
\(
    H(\mathcal{D})
    =
    \frac{1}{N_s}
    \sum_{i=1}^{N_s}H(I_i).
\)
Higher values indicate that individual images exhibit more diverse
pixel-intensity distributions and therefore greater heterogeneity. 


\vspace{0.5em}
\textit{\textbf{Local intrinsic-complexity score.}}
%
%
Let $\mathcal D_i$ denote the local dataset of client $i$, with
$H_i=H(\mathcal D_i)$,
$C_{S,i}=C_S(\mathcal D_i)$, and
$\operatorname{ID}_i=\widehat{\operatorname{ID}}(\mathcal D_i)$.
Any of these indicators can be used directly as the local intrinsic-complexity score $c_i$ without rescaling; but when multiple indicators are combined, they must first be placed on comparable numerical scales.
Denoting the rescaled values by
$\widetilde H_i$, $\widetilde C_{S,i}$, and
$\widetilde{\operatorname{ID}}_i$, we define
\begin{equation}
    c_i
    =
    w_H\widetilde H_i
    +
    w_S\widetilde C_{S,i}
    +
    w_{\mathrm{ID}}\widetilde{\operatorname{ID}}_i,
    \label{eq:local-intrinsic-score}
\end{equation}
where $w_H,w_S,w_{\mathrm{ID}}\geq0$ and
$w_H+w_S+w_{\mathrm{ID}}=1$. 

All three indicators are computed directly from the local data
before model training and do not require class labels, classifier
metrics, or training dynamics. 


\subsection{\textbf{Federated Learning Complexity}}
\label{sec:federated-complexity}

In federated learning, data are partitioned across participating clients, and learning proceeds through iterative communication rounds. 
The client data composition and 
the fragmentation of local data-distribution profiles
can substantially influence learning behavior, even when the underlying model and optimization procedure remain fixed.

In this work, we use federated learning as a representative distributed training paradigm to model and analyze the impact of data composition on learning difficulty. Rather than modifying the federated learning process or proposing a new training strategy, we focus on characterizing how intrinsic data complexity interacts with distributed environment properties to influence task difficulty.


Equation~\eqref{eq:local-intrinsic-score}
defines an intrinsic-complexity score
$c_i$ for the local dataset of each participating client. We now
aggregate these local scores across the federated environment.
For $N_c$ clients in the federated configuration, let
$\mathbf{c}=[c_1,\ldots,c_{N_c}]$ denote the vector of client-level
intrinsic-complexity scores. 
Then the \textbf{federated intrinsic-complexity} is defined as
\begin{equation}
    f(\mathbf{c})
    =
    \beta\lVert \mathbf{c} \rVert_2
    =
    \beta\sqrt{\sum_{i=1}^{N_c}c_i^2},
    \label{eq:federated-intrinsic}
\end{equation}
where $\beta>0$ controls the scale of the aggregation.
%
Equation~\eqref{eq:local-intrinsic-score} combines complementary intrinsic properties
within client $i$, whereas equation~\eqref{eq:federated-intrinsic} aggregates the resulting local scores across participating clients.

The term $f(\mathbf{c})$ characterizes the intrinsic properties of the
participating clients' data, but it does not characterize how those
clients are distributed among distinct data-distribution types.


%
%
Let $d$ denote the number of distinct client-distribution types,
and let \(\mathbf m=[m_1,\ldots,m_d]\)
denote the corresponding type-frequency vector, where $m_j$ is
the number of clients assigned to distribution type $j$ such that
\(\sum_{j=1}^{d}m_j=N_c\), for \(N_c\) clients in the federated configuration. 
To characterize the complexity introduced by data distribution across entities, we define the \textbf{distributed-environment complexity} as:

\begin{equation}
f(d; \mathbf m)
    =
\sum_{j=1}^{d}\frac{1}{m_j}. 
\label{eq:distributed-env-complexity}
\end{equation}

%

For a federated configuration, clients are grouped according to their local label-support profiles:
two clients belong to the same distribution type when their local
datasets contain the same set of class labels. Dividing an existing
type into smaller, distinct groups increases \fdist;
therefore, larger values represent greater fragmentation of client
data distributions. This quantity characterizes client composition
and does not represent round-level participation frequency.

%
Combining equations \eqref{eq:federated-intrinsic} and \eqref{eq:distributed-env-complexity} yields the \textbf{federated learning complexity} metric\footnote{Because each experimental value of $d$ is associated with one fixed
type-frequency vector $\mathbf m(d)$, we use the shorthand $f(d)$ and $F(d, \mathbf{c})$
in the results and discussion.} 
for a given federated configuration:
\begin{equation}
    F(d, \mathbf{c}; \mathbf{m})
    =
    \alpha_c f(\mathbf{c})
    +
    \alpha_d f(d; \mathbf{m}),
    \label{eqn:fdx}
\end{equation}
where $\alpha_c$ and $\alpha_d$ control the relative contributions
of the intrinsic and distributed terms. In this work, we consider $\alpha_c$ = $\alpha_d$ = 1.


For the heterogeneity-only instantiation used in the controlled
experiments, we set
\(
w_H=1
\)
and
\(
w_S=w_{\mathrm{ID}}=0
\),
so that
\(
c_i=H(\mathcal{D}_i)
\).
We also set \(\beta=1\). To keep the intrinsic term fixed while varying
client-distribution composition, we approximate each local
heterogeneity using the complete-training-set value:
\(
H(\mathcal{D}_i)\approx H(\mathcal{D})
\).
Consequently,
\begin{equation}
\begin{aligned}
    f(\mathbf{c})
    &=
    \sqrt{
        \sum_{i=1}^{N_c}
        H(\mathcal{D}_i)^2
    }\\
    &\approx
    \sqrt{N_c}\,H(\mathcal{D})
    \equiv
    f_{\mathrm{proxy}}(\mathbf{c}).
    \label{eqn:experiment-fdx}
\end{aligned}
\end{equation}

The accuracy of this approximation is evaluated using direct
client-level measurements in
Table~\ref{tab:client_heterogeneity}.
Accordingly, the metric evaluated in
the controlled comparisons is
\begin{equation}
    F(d,\mathbf{c};\mathbf{m})
    =
    f_{\mathrm{proxy}}(\mathbf{c})
    +
    f(d;\mathbf{m}).
\end{equation}

The proposed metric is used as a pre-training diagnostic of expected
learning difficulty. It enables federated configurations to be
compared in terms of anticipated accuracy and communication effort
without changing the learning algorithm or training method. 


\section{EXPERIMENTAL SETUP} 





The primary federated-learning experiments were conducted using Python 3.8 on an Intel Core i7 system with 16 GB of memory, while subsequent validation experiments used Python 3.9 in GPU-accelerated environments. Standard open-source libraries were used throughout, with Keras-based implementations for the primary experiments and PyTorch-based implementations for the subsequent analyses.

\textbf{Datasets.}
We evaluate the proposed framework using three commonly used perception benchmarks: Handwritten-MNIST \cite{deng2012mnist}, Fashion-MNIST \cite{xiao2017fashionmnist}, and EMNIST-Digits \cite{emnist2017}. These datasets provide controlled variations in visual complexity while maintaining identical input resolution and label structure, making them suitable for comparative complexity analysis.

Handwritten-MNIST and Fashion-MNIST each contain 70{,}000 grayscale images, with 60{,}000 training samples and 10{,}000 test samples. EMNIST-Digits contains 280{,}000 digit images. All images are of size $28 \times 28$ pixels with pixel intensities in the range $[0,255]$.




\vspace{0.5em}
\textbf{Intrinsic Complexity Measurements.}
All intrinsic measurements are computed from the training split
before model training. The datasets are loaded using Torchvision
\cite{paszke2019pytorch} and evaluated on their original grayscale
intensity scale $[0,255]$. Mean per-image Shannon entropy
\cite{shannon1948mathematical} and Hoyer sparsity
\cite{hoyer2004nonnegative} are implemented using NumPy
\cite{harris2020array}. Entropy is calculated from the 256-bin
intensity histogram of each image using base-two logarithms, and the
resulting image-level values are averaged to obtain
$H(\mathcal D)$. Hoyer sparsity is calculated from each flattened
image and averaged to obtain $S(\mathcal D)$, from which
$C_S(\mathcal D)=1-S(\mathcal D)$ is obtained.

The variance-filtered ambient feature count is computed using
Scikit-learn's \texttt{VarianceThreshold}
\cite{scikit-learn} with \texttt{threshold=90} on flattened
images in the original pixel scale. Intrinsic dimensionality is
estimated using the maximum-likelihood estimator of Levina and
Bickel \cite{levina2004maximum}, as implemented by
Scikit-dimension \cite{bac2021scikit}. 
The estimator is fitted using 20 nearest neighbors.
Table~\ref{tab:dataset_measurements} summarizes the resulting measurements.

\subsection{\textbf{Federated Learning Configuration}}
Federated learning experiments are conducted using a fixed shallow convolutional neural network (CNN) architecture across all settings.
For MNIST variants, we used shallow ProbeNet variant \cite{Scheidegger2021} as our fixed CNN evaluation model. 
Figure~\ref{fig:relationship} shows broad agreement between the dataset-level difficulty indicated by ProbeNet accuracy and that indicated by the published benchmark accuracies for the three datasets \cite{an2020ensemble, tanveer2021fine,viswanathanwavemix}. Both ProbeNet and the published benchmark results
distinguish Fashion-MNIST as substantially more difficult than Handwritten-MNIST and EMNIST-Digits, although the exact ordering of the two digit datasets differs very slightly. This positive association supports the use of ProbeNet for controlled comparisons of federated-learning behavior across the three datasets. 
Throughout the remainder of the paper, we use ProbeNet accuracy as an inverse empirical indicator of learning difficulty: lower accuracy indicates a more difficult learning task. ProbeNet is used only as the evaluation model for obtaining federated-learning outcomes against which the diagnostic value of the proposed complexity metric is assessed. The proposed complexity measures remain classifier-agnostic.
%

All federated clients use the same shallow ProbeNet architecture and setting \cite{Scheidegger2021}  as local training procedure, while the server performs aggregation using FedAvg \cite{44822}. 
Across the experiments, the federated environment contains five clients with comparable local data volumes. Two of the five clients participate in each communication round, and training proceeds for 100 communication rounds.
In our experiments, \(N_c=5\), and the five configurations divide the clients into \(d\in\{1,\ldots,5\}\) distinct distribution types, with type-frequency vectors \((5)\), \((3,2)\), \((2,2,1)\), \((2,1,1,1)\), and \((1,1,1,1,1)\), respectively. Here, for example, when $d=2$, the type-frequency vector is
$\mathbf{m} = (3,2)$, indicating that three clients share one
distribution type and two clients share another. Therefore,
\(
f(d)
=\frac{1}{3}+\frac{1}{2}
= \frac{5}{6}
\approx0.833. 
\)
These configurations produce \(f(d)=0.20,\ 0.833,\ 2.00,\ 3.50,\) and \(5.00\).
%
Client datasets are non-IID but comparably sized; clients assigned to the same distribution type may contain different samples while satisfying the same distribution-type criterion. This design limits data-volume disparity while varying client-distribution fragmentation.
To assess whether the combined complexity relationship depends on a particular client-support realization, we additionally evaluated deterministic client-support configurations preserving the same type-frequency structure, client count, participation rate, and common federated protocol. In these experiments, the complete training split is distributed across clients without duplication or omission, and each selected client performs one full local epoch. The  implementation uses RMSProp with learning rate \(10^{-4}\), batch size 32, deterministic client selection schedule initialized from seed 42, and sample-count-weighted FedAvg, with native \(28\times28\) grayscale inputs and no data augmentation.
Separately, to evaluate the fixed global approximation in Table~\ref{tab:client_heterogeneity}, we measure
client-level heterogeneity under the representative
\(d=3,\ f(d)=2\) configuration, whose type-frequency vector is
\(\mathbf{m}=(2,2,1)\). The five client label-support sets are
\(
\mathcal{L}_1=\mathcal{L}_2=\{0,1,2,3\},
\mathcal{L}_3=\mathcal{L}_4=\{4,5,6\},
\mathcal{L}_5=\{7,8,9\}.
\)
Each client contains \(6{,}000\) training images. Classes are
represented uniformly within each client's label-support set, and
individual images are sampled randomly without replacement using a
fixed seed of 42. 
%
For each client, \(H(\mathcal{D}_i)\) is computed from its local images and aggregated according to Equation~(6), with \(\beta=1\), to obtain the measured \(f(\mathbf{c})\). 
We compare this value against
\(
f_{\mathrm{proxy}}(\mathbf{c})=\sqrt{5}H(\mathcal{D})
\)
to assess whether the complete-dataset heterogeneity provides an
adequate fixed approximation of client-level intrinsic complexity.

\vspace{0.5em}
\textbf{Evaluation Metrics.}
We evaluate federated learning behavior using two observable quantities:
\emph{accuracy} and \emph{communication effort}. Accuracy is reported as both the maximum test accuracy attained during training and the average test accuracy across communication rounds. Communication effort is the number of rounds required to reach the fixed accuracy threshold of 60\%.

We assess the diagnostic value of the proposed framework by examining the relationships between these outcomes and the distributed term \(f(d;\mathbf{m})\), the intrinsic term \(f(\mathbf{c})\), and their combined value \(F(d,\mathbf{c};\mathbf{m})\). 

\section{RESULTS AND ANALYSIS}


\subsection{\textbf{Intrinsic Complexity Across Datasets}} 

\begin{table*}[h]
\centering
\caption{Dataset-level heterogeneity, sparsity, and dimensionality
measurements for the MNIST variants.}
\label{tab:dataset_measurements}
\begin{tabular}{l|c|cc|cc}
\hline
& \multicolumn{1}{c|}{Heterogeneity}
& \multicolumn{2}{c|}{Sparsity}
& \multicolumn{2}{c}{Dimensionality} \\
\cline{2-2}
\cline{3-4}
\cline{5-6}
Dataset
& \(H(\mathcal{D})\)
& \(S(\mathcal{D})\)
& \(C_S(\mathcal{D})\)
& \(ID(\mathcal{D})\)
& \(N_f^r(90)\) \\
\hline
Handwritten-MNIST & 1.6025 & 0.6365 & 0.3635 & 13.368 & 530 \\
EMNIST-Digits     & 2.8692 & 0.5607 & 0.4393 & 14.095 & 557 \\
Fashion-MNIST     & 4.1165 & 0.3798 & 0.6202 & 14.547 & 745 \\
\hline
\end{tabular}
\end{table*}





\begin{table*}[t]
\centering
\caption{Comparison of the fixed global heterogeneity proxy and the
measured client-level intrinsic-complexity aggregation under the
representative \(d=3,\ f(d)=2\) configuration. Client heterogeneity is
reported as the mean and standard deviation across the five
client-level values.} 
\label{tab:client_heterogeneity}
\begin{tabular}{lccccc}
\hline
Dataset
& \(H(\mathcal{D})\)
& Client \(H(\mathcal{D}_i)\)
& \(f_{\mathrm{proxy}}(\mathbf{c})\)
& Measured \(f(\mathbf{c})\)
& Relative difference \\
\hline

Handwritten-MNIST
& 1.6025
& \(1.6130\pm0.0186\)
& 3.5833
& 3.6071
& \(+0.66\%\) \\

EMNIST-Digits
& 2.8692
& \(2.8678\pm0.0287\)
& 6.4157
& 6.4128
& \(-0.05\%\) \\

Fashion-MNIST
& 4.1165
& \(4.1377\pm0.0812\)
& 9.2048
& 9.2540
& \(+0.53\%\) \\

\hline
\end{tabular}
\end{table*}

Table~\ref{tab:dataset_measurements} reports the pre-training intrinsic complexity
measurements for the three datasets, calculated on the complete training split before client partitioning. The measurements consistently
place Handwritten-MNIST, EMNIST-Digits, and Fashion-MNIST in increasing
order of heterogeneity, sparsity complexity, and dimensionality, reinforcing their use
as representative lower-, intermediate-, and higher-complexity cases in the subsequent experiments.

The subsequent controlled experiments instantiate the federated intrinsic-complexity term using heterogeneity; sparsity and dimensionality are reported as complementary intrinsic diagnostics. 

\paragraph{\textbf{Validation of the Global Heterogeneity Proxy.}}
Table~\ref{tab:client_heterogeneity} compares the fixed global
heterogeneity proxy
\(
f_{\mathrm{proxy}}(\mathbf{c})=\sqrt{5}H(\mathcal{D})
\)
with the intrinsic-complexity value
\(
f(\mathbf{c})
\)
calculated directly from the five client-level heterogeneity
measurements. The client column reports the mean and standard deviation
of
\(
\{H(\mathcal{D}_i)\}_{i=1}^{5}
\),
and hence describes variation among client-level mean entropy
values rather than variation among individual images.

The five client-level values exhibit limited dispersion within each
dataset. The relative differences are
\(0.66\%\) for Handwritten-MNIST, \(0.05\%\) for EMNIST-Digits, and
\(0.53\%\) for Fashion-MNIST, with a mean absolute relative difference
of \(0.41\%\). More importantly, the resulting \(f(\mathbf{c})\) closely
agrees with \(f_{\mathrm{proxy}}(\mathbf{c})\), supporting its use as a fixed approximation in the controlled experiments. 



\subsection{\textbf{Distributed Environment Complexity}} 

\begin{figure*}
	\centering
	\begin{minipage}{.65\columnwidth}
		\centering
		\includegraphics[width=1\textwidth]{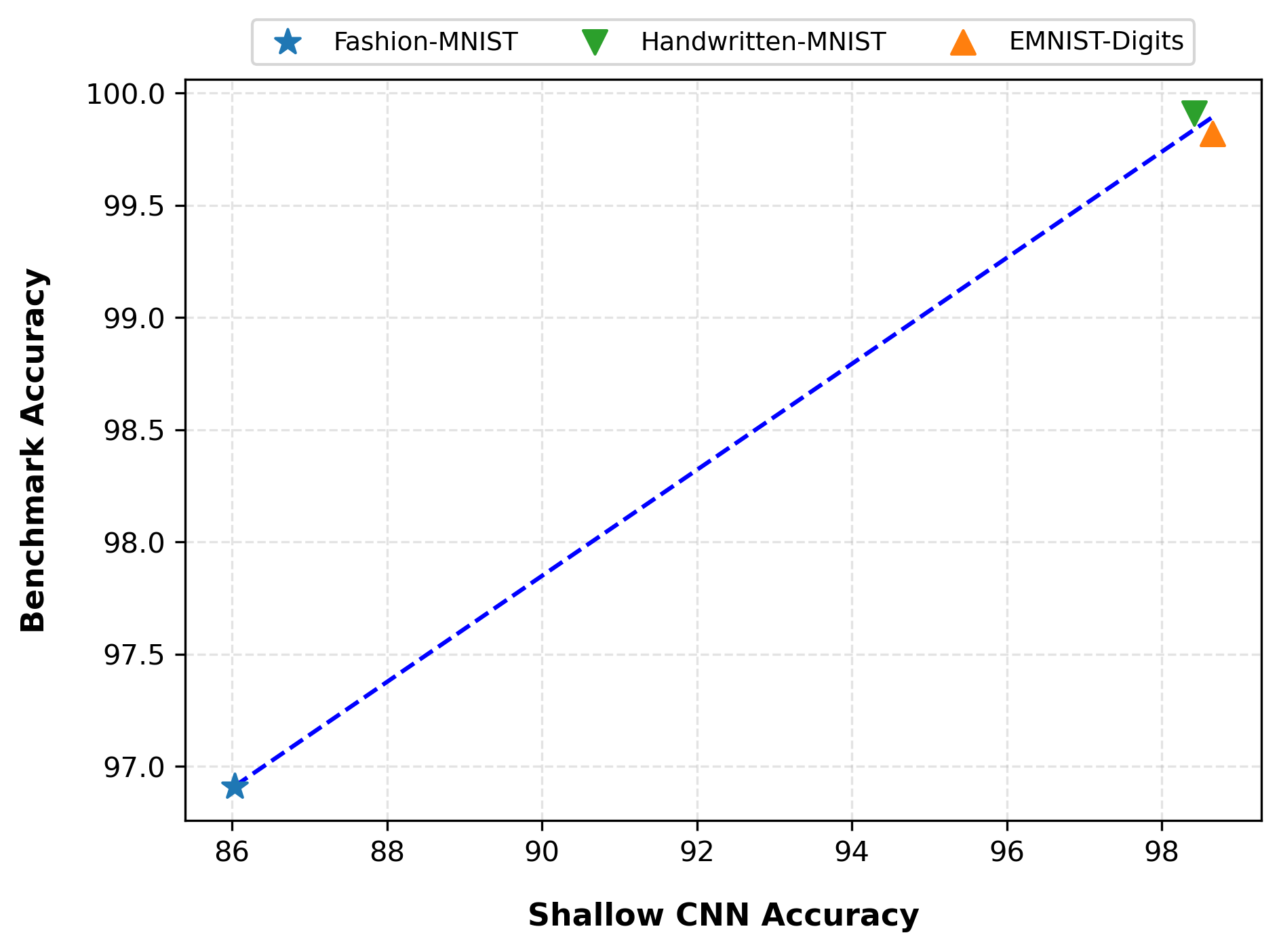}
        \subcaption{}
		\label{fig:relationship}
	\end{minipage}%
	\begin{minipage}{.65\columnwidth}
		\centering
		\includegraphics[width=1\textwidth]{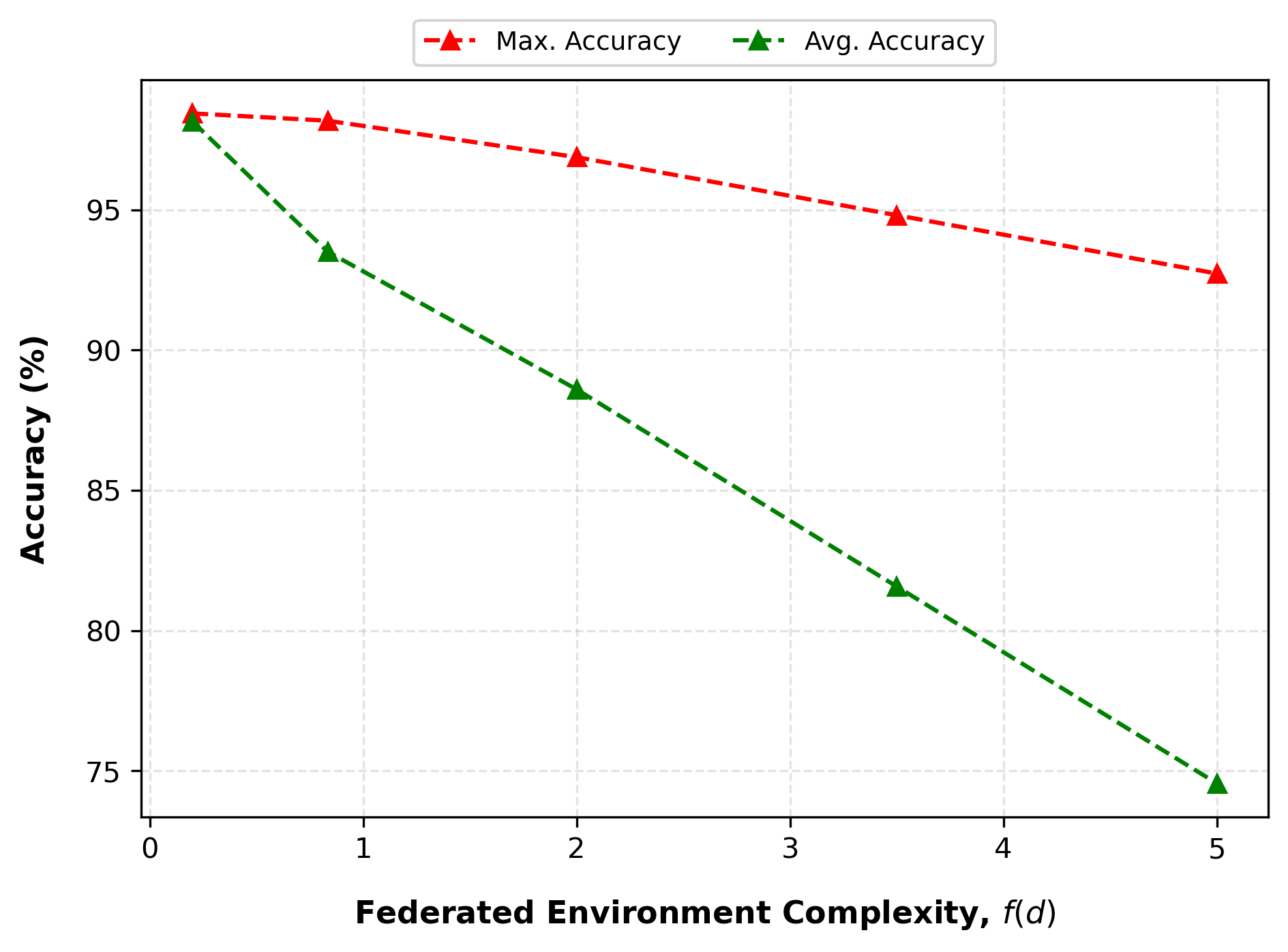}
        \subcaption{}
		\label{fig:rel1}
	\end{minipage}
	\begin{minipage}{.65\columnwidth}
		\centering
		\includegraphics[width=1\textwidth]{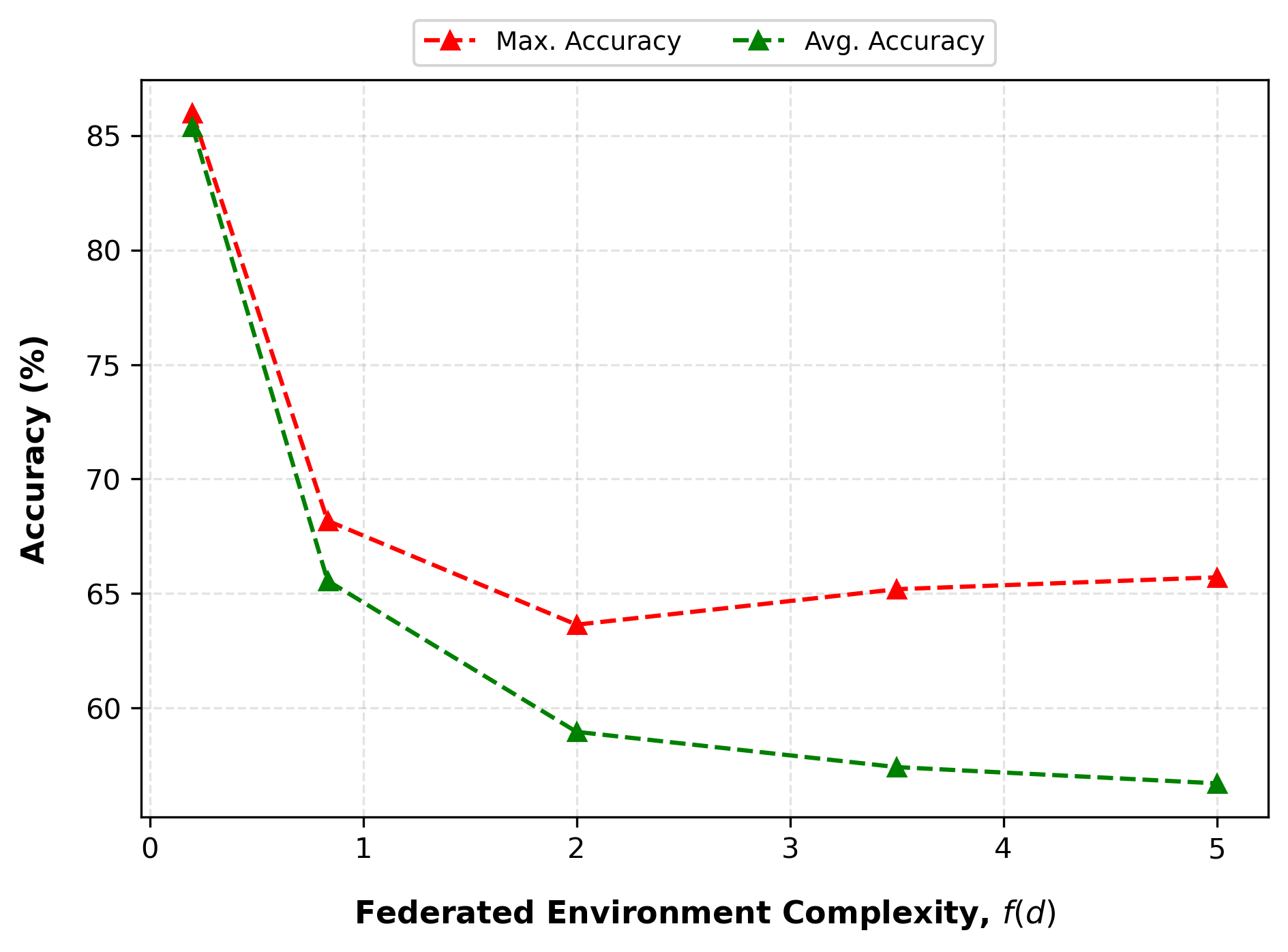}
        \subcaption{}
		\label{fig:rel3}
	\end{minipage}
        \caption{(a) Shallow CNN accuracy vs. published benchmark accuracy for Fashion-MNIST (96.91\%, fine-tuned DARTS \cite{tanveer2021fine}), Handwritten-MNIST (99.91\%, CNN ensemble \cite{an2020ensemble}) and EMNIST-Digits (99.82\%, WaveMix \cite{viswanathanwavemix}), (b) Federated environment complexity $f(d)$ vs. federated-learning accuracy for Handwritten-MNIST with $f(\mathbf{c})$ fixed, (c) Federated environment complexity $f(d)$ vs. federated-learning accuracy for Fashion-MNIST with $f(\mathbf{c})$ fixed. }
        \label{fig-master-2}
\end{figure*}

Figures~\ref{fig:rel1} and~\ref{fig:rel3} illustrate the relationship between federated environment complexity \fdist\ and federated learning accuracy for Handwritten-MNIST and Fashion-MNIST, respectively, with intrinsic complexity held fixed. 
As the number of distinct client distribution types increases from one to five while the total number of clients remains fixed at \(N_c=5\), client-distribution fragmentation and \fdist\ increase.
For both datasets, maximum and average accuracy generally decrease as \fdist\ increases, with a stronger decline in average accuracy. For Fashion-MNIST, maximum accuracy declines sharply at lower \fdist\ values and then stabilizes with a small recovery at higher values.

\begin{figure*}
	\centering
	\begin{minipage}{.65\columnwidth}
		\centering
		\includegraphics[width=1.0\textwidth]{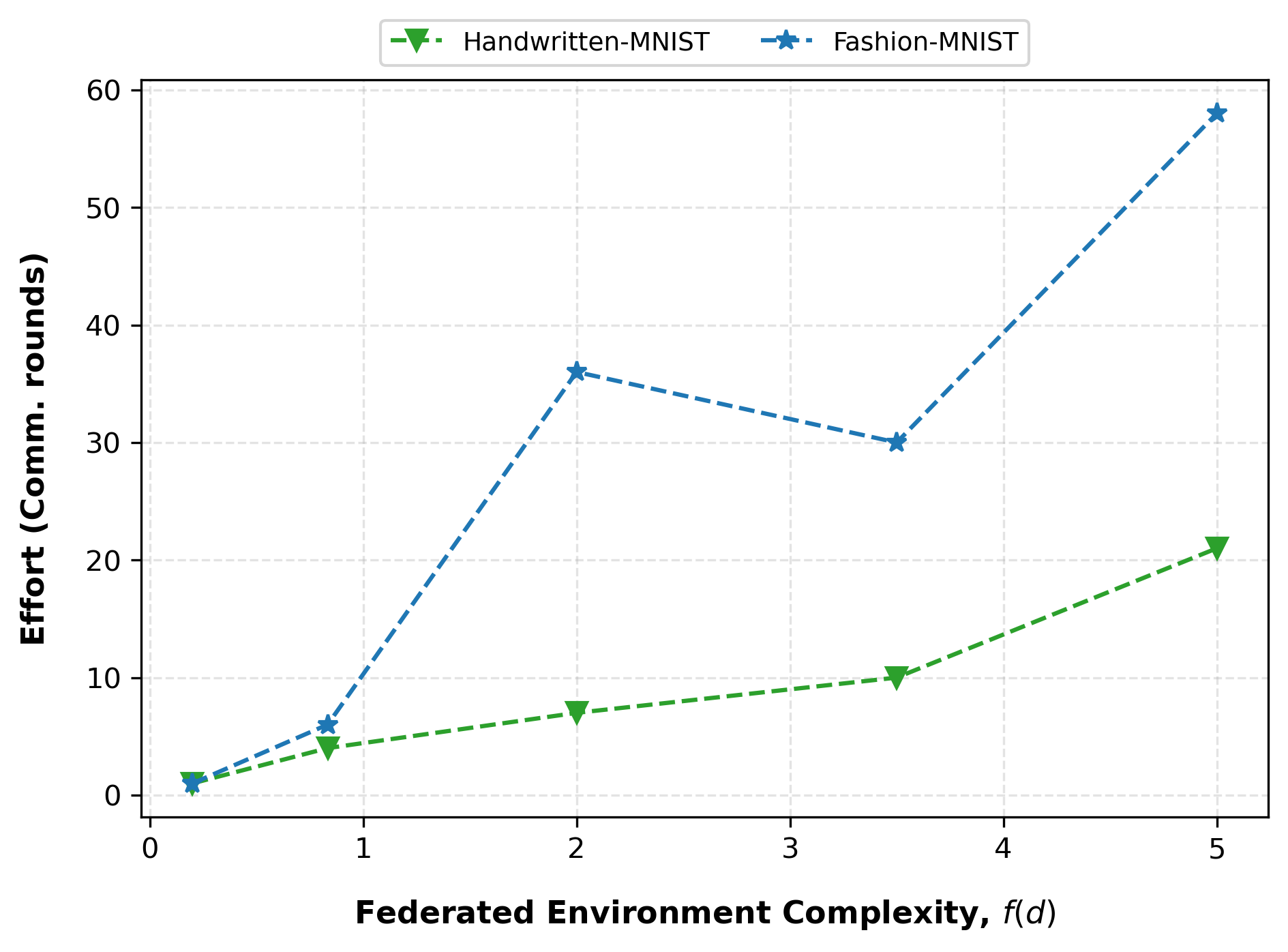}
        \subcaption{}
		\label{fig:rel2}
	\end{minipage}%
	\begin{minipage}{.65\columnwidth}
		\centering
		\includegraphics[width=1.0\textwidth]{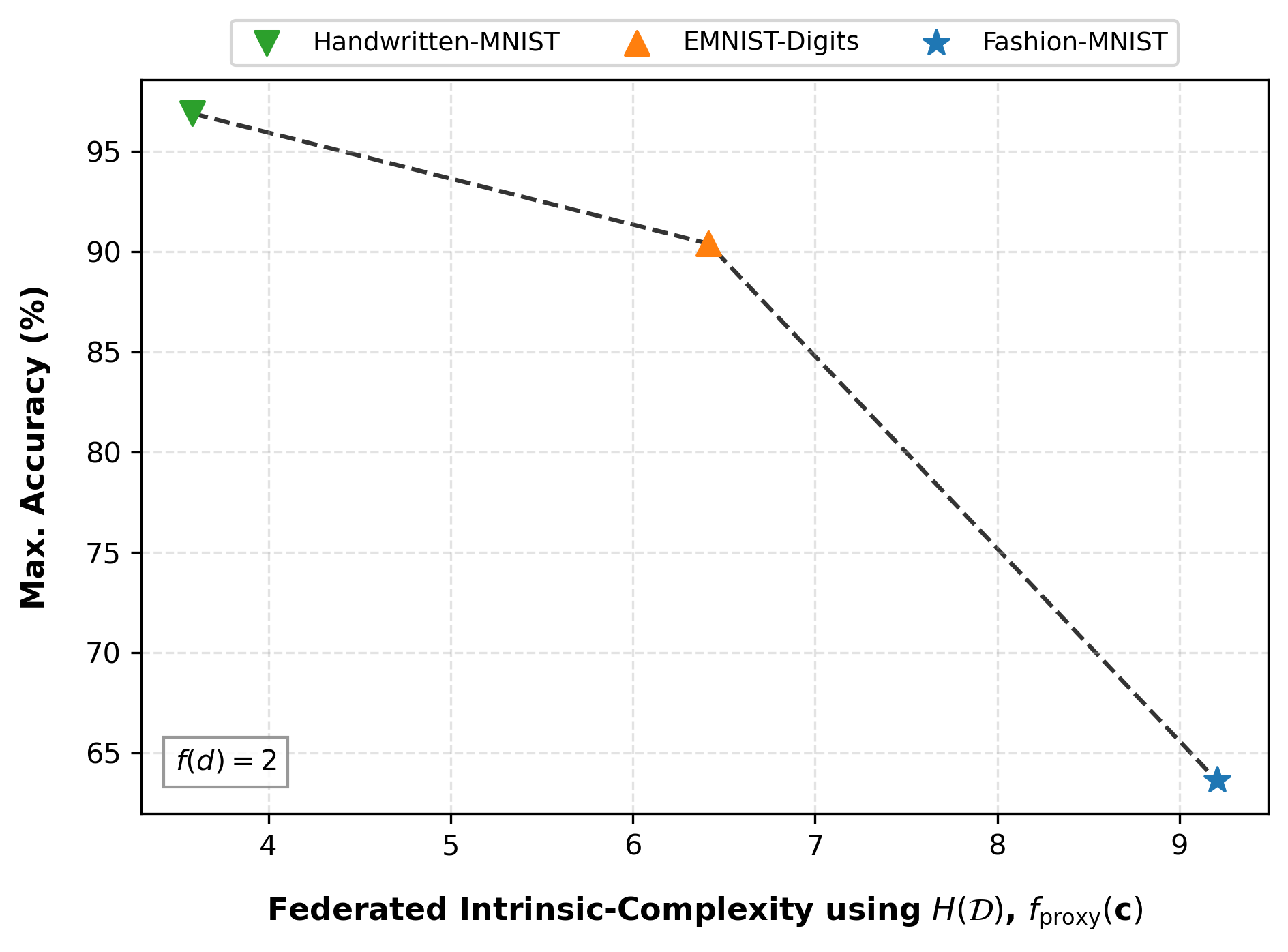}
		\subcaption{}
		\label{fig:intrinsic-vs-max-acc}
	\end{minipage}
	\begin{minipage}{.65\columnwidth}
		\centering
		\includegraphics[width=1.0\textwidth]{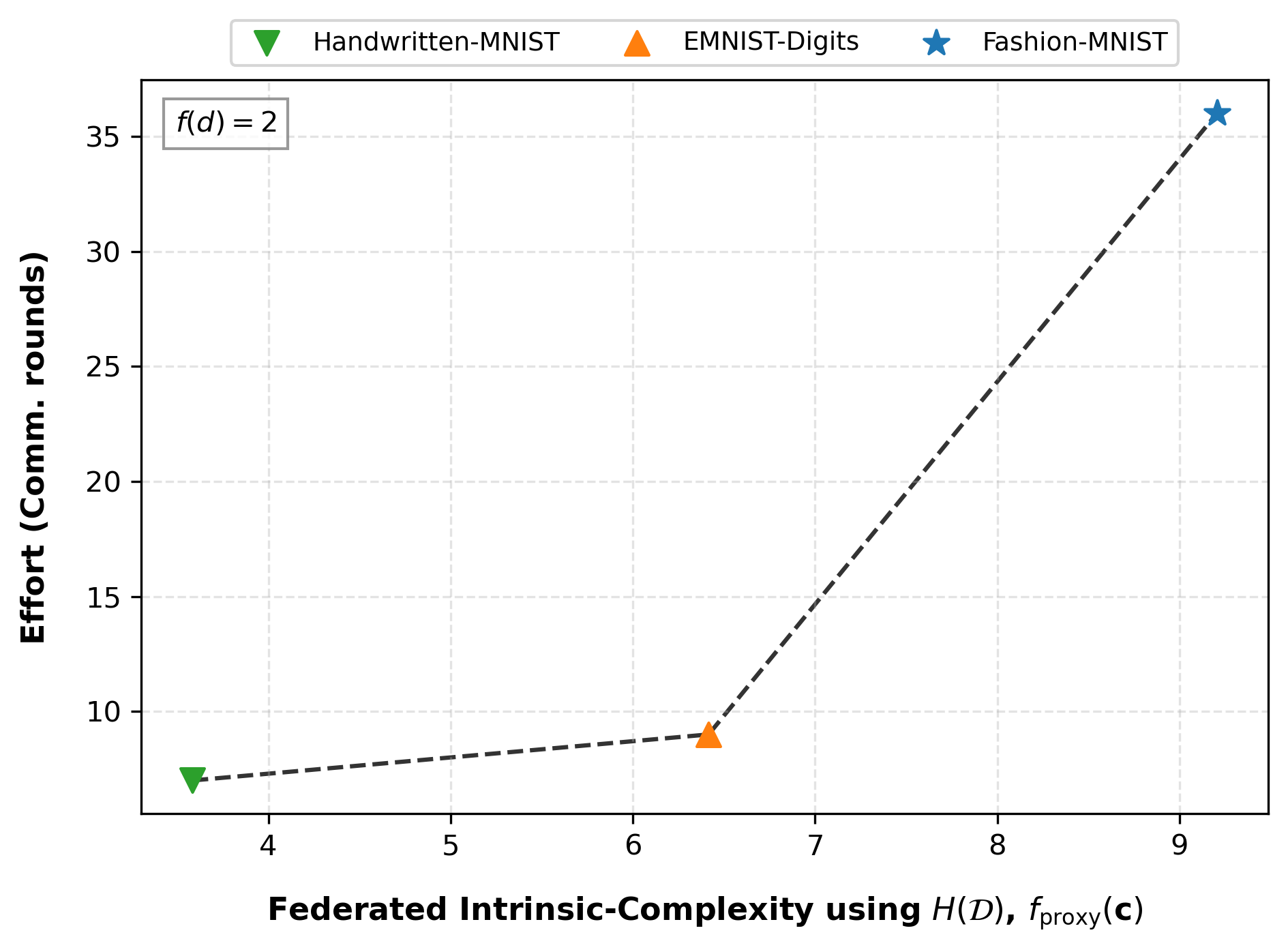}
		\subcaption{}
		\label{fig:intrinsic-vs-effort}
	\end{minipage}
        \caption{(a) Federated environment complexity $f(d)$ vs. communication effort
        for Handwritten-MNIST and Fashion-MNIST; (b)  heterogeneity-based federated intrinsic-complexity \(f_{\mathrm{proxy}}(\mathbf c)\) vs. maximum federated-learning accuracy 
        across MNIST-variants, with $f(d)= 2$; (c) 
        \(f_{\mathrm{proxy}}(\mathbf c)\) vs. communication effort across MNIST-variants with $f(d) = 2$.
        }
        \label{fig-master-3}
\end{figure*}

Figure~\ref{fig:rel2} further shows that 
communication effort generally increases with distributed-environment complexity. The relationship is approximately monotonic with \fdist\ for Handwritten-MNIST, whereas Fashion-MNIST exhibits an intermediate fluctuation but a clear overall increase between the lowest- and highest-complexity configurations.
Collectively, these results suggest that greater distributed-environment complexity is generally associated with reduced accuracy and increased communication effort.

Only lower- and higher-complexity representative cases are shown here; all three datasets are included in subsequent analysis. 

\vspace{0.3em}
\textit{\textbf{Impact of Intrinsic Complexity on Federated Learning.}}
Figures~\ref{fig:intrinsic-vs-max-acc} and \ref{fig:intrinsic-vs-effort} examine the relationship between the heterogeneity-based intrinsic-complexity proxy \(f_{\mathrm{proxy}}(\mathbf{c})\) and federated learning behavior while holding the distributed-environment complexity fixed at \(f(d)=2\). Across the three datasets, higher \(f_{\mathrm{proxy}}(\mathbf{c})\) is associated with lower maximum accuracy and greater communication effort. Handwritten-MNIST has the lowest intrinsic-complexity proxy and the highest accuracy and lowest effort, whereas Fashion-MNIST has the highest proxy value and exhibits the lowest accuracy and greatest effort. EMNIST-Digits lies between these extremes, yielding a consistent three-dataset ordering between heterogeneity-based intrinsic complexity and learning difficulty.

\subsection{\textbf{Federated Learning Complexity}}

\begin{figure*}[htbp]
    \centering
    \begin{minipage}[t]{0.48\textwidth}
        \centering
        \includegraphics[width=\linewidth]{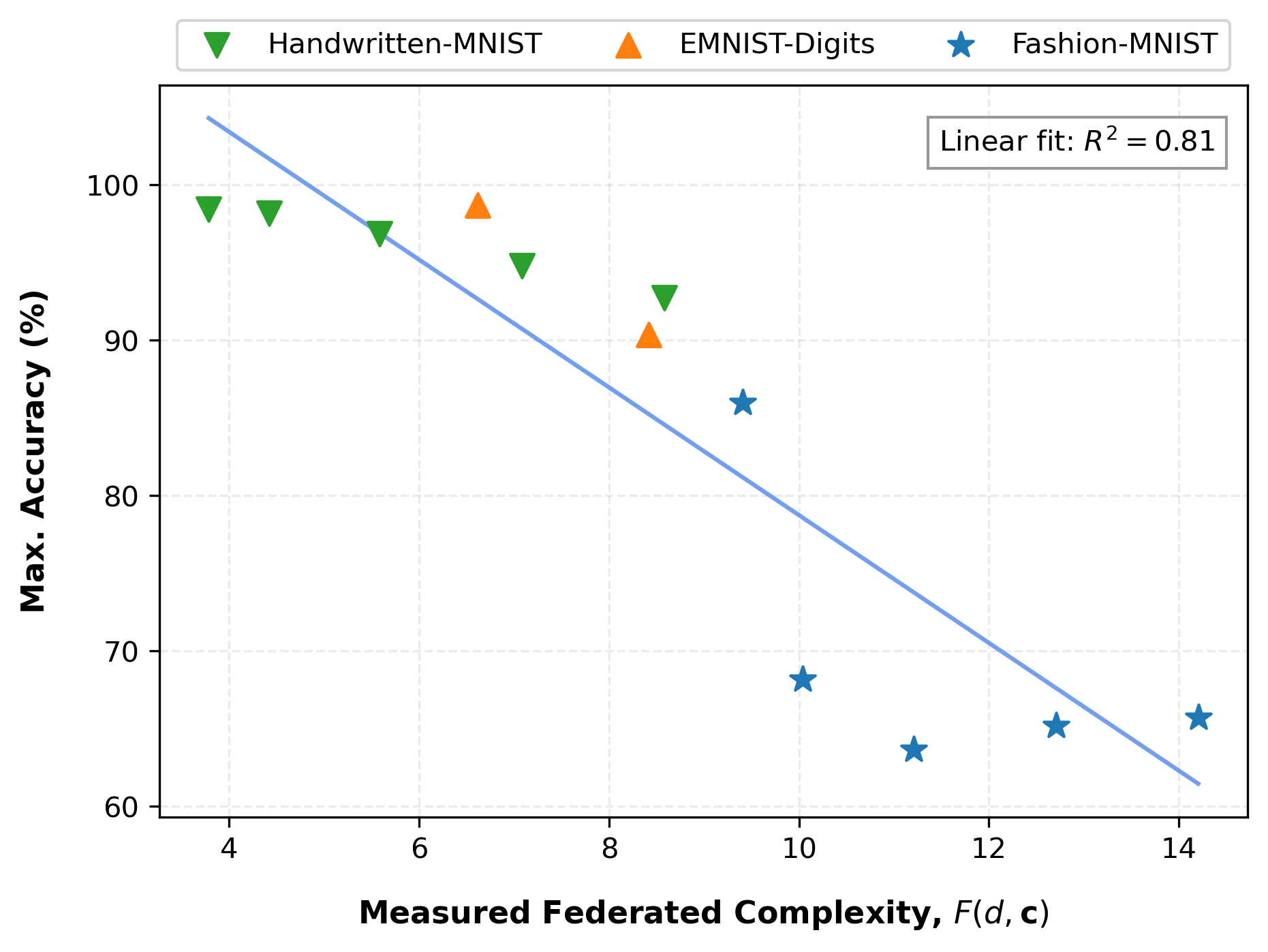}
        \subcaption{maximum accuracy (\(R^2=0.81\))}
        \label{fig:fdx-vs-max-acc}
    \end{minipage}
    \hfill
    \begin{minipage}[t]{0.48\textwidth}
        \centering
        \includegraphics[width=\linewidth]{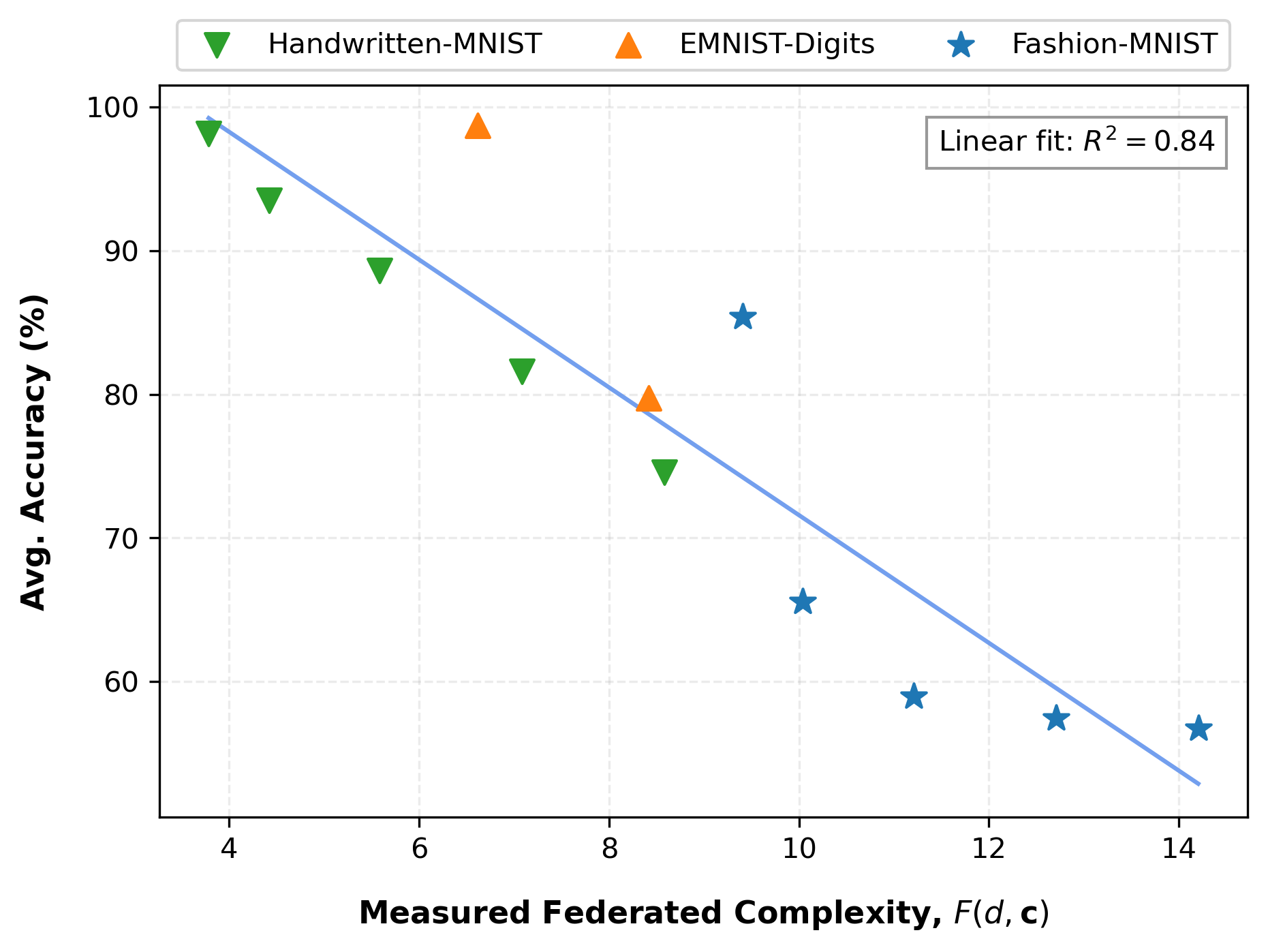}
        \subcaption{average accuracy (\(R^2=0.84\))}
        \label{fig:fdx-vs-avg-acc}
    \end{minipage}

    \caption{Federated complexity \(F(d,\mathbf{c})\) vs. shallow federated-learning accuracy across MNIST-variant datasets.
    Higher estimated federated complexity is associated with lower accuracy.}
    \label{fig:fdx-vs-acc}
\end{figure*}

Figures~\ref{fig:fdx-vs-max-acc} and \ref{fig:fdx-vs-avg-acc} evaluate the relationship between the combined federated learning complexity metric
\(
    F(d,\mathbf{c})
    =
    f_{\mathrm{proxy}}(\mathbf{c})+f(d)
\)
and federated learning accuracy. 
The analysis includes 12 observations: five Handwritten-MNIST client-distribution configurations, five Fashion-MNIST configurations, and two EMNIST-Digits configurations (\(d=1\) and \(d=3\)).
The combined metric exhibits a strong negative correlation with maximum accuracy
\(
(r=-0.900,\ R^2=0.810)
\)
and average accuracy
\(
(r=-0.919,\ R^2=0.845)
\).
Thus, configurations with greater estimated federated learning complexity generally achieve lower accuracy.
%
The slightly stronger correlation with average accuracy suggests that \(F(d,\mathbf{c})\) more closely reflects sustained learning behavior across communication rounds than isolated peak performance. 

The deterministic validation configurations reproduced the same qualitative accuracy trends observed in Figures \ref{fig:rel1} and \ref{fig:rel3}. When the combined complexity metric was evaluated across the Handwritten-MNIST and Fashion-MNIST validation configurations, the resulting relationships remained comparably strong (\(R^2 \approx 0.823\) for maximum accuracy and \(R^2 \approx 0.845\) for average accuracy), providing an additional robustness check on the combined-complexity relationship shown in Figure \ref{fig:fdx-vs-acc}.
For comparison, the distributed term \(f(d)\) alone yields \(R^2=0.111\) and \(0.391\) for maximum and average accuracy, respectively, while the intrinsic term \(f_{\mathrm{proxy}}(\mathbf{c})\) yields \(R^2=0.737\) and \(0.473\). The combined metric \(F(d,\mathbf{c})\) provides the strongest overall association, with \(R^2=0.810\) and \(0.845\), respectively.
These results support use of \(F(d,\mathbf{c})\) as a pre-training diagnostic for comparing the expected learning difficulty of federated configurations.




\section{DISCUSSION AND FUTURE WORK}


A key implication of this work is that federated learning difficulty is not determined solely by the choice of model architecture or optimization algorithm. Even when training configurations are held constant, variations in dataset structure and client composition lead to measurable differences in convergence behavior and communication effort. The proposed framework makes these differences explicit prior to training, enabling comparison of federated configurations without requiring extensive trial-and-error experimentation.

From an edge AI perspective, such pre-deployment complexity estimation is particularly valuable. Edge-deployed systems often operate under strict resource constraints, where communication budgets, energy consumption, and latency must be considered alongside accuracy. By providing an early indication of expected learning difficulty, the proposed metric can support feasibility assessment, dataset selection, and resource planning before committing to large-scale federated training.

The results provide evidence that federated learning behavior is associated with both intrinsic data properties and client-distribution composition. With the intrinsic term held fixed, increasing distributed-environment complexity \(f(d)\) is generally associated with lower accuracy and greater communication effort. Conversely, with a fixed \(f(d)\), the intrinsic-complexity \(f_{\mathrm{proxy}}(\mathbf{c})\) orders the three datasets consistently with their observed accuracy and communication effort. Combining these terms yields strong negative correlations with accuracy across the evaluated configurations.


%
The proposed metric is intended as a diagnostic rather than an optimization mechanism. It characterizes task difficulty independently of training dynamics and complements existing federated optimization methods without prescribing client selection, training schedules, or model adaptations.


While these controlled settings support evaluation of the diagnostic relationship between estimated complexity and learning behavior, they do not capture the full diversity of real-world federated systems. The present experiments use one representative type-frequency vector for each \(d\); evaluating alternative frequency compositions while holding \(d\) fixed remains an important extension. Other factors, including extreme data imbalance, heterogeneous client hardware, variable client participation rates and availability, complex data modalities, and interactions with adaptive training strategies, also remain outside the present scope.


\section{CONCLUSION}

This paper introduced a pre-training diagnostic framework for estimating learning complexity in federated perception systems. 
The framework separates intrinsic data complexity from distributed-environment complexity and combines them into a unified diagnostic score. Intrinsic complexity may be represented using heterogeneity, sparsity-derived complexity, intrinsic dimensionality, or 
with a combination of these indicators. The distributed term characterizes fragmentation among client label-support types.

%
Across 
five client-distribution configurations in a federated training paradigm
spanning three MNIST-variant datasets, the resulting combined metric exhibited strong negative correlations with both maximum and average federated accuracy. Distributed and intrinsic complexity were also individually associated with accuracy and communication effort when the other component was held fixed. These findings provide initial evidence that pre-training complexity estimation can support comparison of expected learning difficulty without modifying the federated learning algorithm. 
Evaluation on larger and more diverse datasets, models, and federated configurations remains an important direction for establishing the broader generality of the framework in real-world edge AI environments.



\section{ACKNOWLEDGMENTS}
This research was supported, in part, by the Defense Advanced Research Projects Agency (DARPA) and the Air Force Research Laboratory (AFRL) under the contract number W911NF2020003. The views and conclusions contained herein are those of the authors and should not be interpreted as necessarily representing the official policies or endorsements, either expressed or implied, of DARPA, AFRL, or the U.S. Government.
KMA Solaiman also acknowledges travel support from the Department of Computer Science and Electrical Engineering at UMBC to present this work at EDGE AI 2026.


\def\refname{REFERENCES}
\bibliographystyle{IEEEtran}
\bibliography{reference}

\input{fedcomplex_appendix_bundle/fedcomplex_appendix}

\end{document}

%% file: fedcomplex_appendix_bundle/fedcomplex_appendix.tex

\appendix

\section*{Deterministic Validation Configurations and Supplementary Results}
\label{app:deterministic-validation}

The main text reports an additional deterministic validation of the
combined-complexity relationship using fully specified client label-support
configurations. This appendix records the exact configurations and the
corresponding accuracy results. The validation retains the type-frequency
vectors used in the controlled study,
\(\mathbf{m}=(5)\), \((3,2)\), \((2,2,1)\), \((2,1,1,1)\), and
\((1,1,1,1,1)\) for \(d=1,\ldots,5\), respectively. Consequently,
\(f(d)=0.2000\), \(0.8333\), \(2.0000\), \(3.5000\), and \(5.0000\).

The validation follows the federated protocol described in the main text:
five clients, two participating clients per communication round, 100 rounds,
one full local epoch per selected client, shallow ProbeNet, and
sample-count-weighted FedAvg. Local optimization uses RMSProp with learning
rate \(10^{-4}\) and batch size 32. The complete official training split is
distributed across the five clients without duplication or omission, client
selection follows the deterministic schedule initialized from seed 42, and
the native \(28\times28\) grayscale inputs are used without augmentation.
Each support shown below is therefore an explicit, reproducible validation
configuration; it is not intended to reconstruct undocumented historical
client assignments.

\subsection{Client Label-Support Configurations}

Tables~\ref{tab:appendix-mnist-supports} and
\ref{tab:appendix-fashion-supports} list the label set available to each
client. Client identifiers are \(C_0,\ldots,C_4\); ranges in subscripts denote
all clients in the range. Labels use the canonical numeric encoding of the
corresponding dataset. Repeated support sets realize the required
type-frequency vector while keeping the total client count fixed at five.

\begin{table*}[t]
\caption{Deterministic Handwritten-MNIST client label-support configurations.}
\label{tab:appendix-mnist-supports}
\centering
\scriptsize
\setlength{\tabcolsep}{4pt}
\renewcommand{\arraystretch}{1.18}
\begin{tabular}{c c c p{0.72\textwidth}}
\hline
\(d\) & \(\mathbf{m}\) & \(f(d)\) & Client label supports \\
\hline
1 & \((5)\) & 0.2000 &
\(C_{0\text{-}4}:\{0,1,2,3,4,5,6,7,8,9\}\) \\
2 & \((3,2)\) & 0.8333 &
\(C_{0\text{-}2}:\{0,1,2,3,4,6,7,8,9\}\); \(C_{3\text{-}4}:\{0,2,3,4,5,6,7,8,9\}\) \\
3 & \((2,2,1)\) & 2.0000 &
\(C_{0\text{-}1}:\{0,1,2,3,4,5,6,7\}\); \(C_{2\text{-}3}:\{2,3,4,5,6,7,8,9\}\); \(C_4:\{0,1,4,5,6,7,8,9\}\) \\
4 & \((2,1,1,1)\) & 3.5000 &
\(C_{0\text{-}1}:\{0,1,3,4,5,7,9\}\); \(C_2:\{2,3,4,5,6,8,9\}\); \(C_3:\{0,1,2,3,6,8,9\}\); \(C_4:\{1,2,3,6,7,8,9\}\) \\
5 & \((1,1,1,1,1)\) & 5.0000 &
\(C_0:\{2,4,5,6,7,8\}\); \(C_1:\{0,1,3,4,5,8\}\); \(C_2:\{0,3,4,7,8,9\}\); \(C_3:\{1,2,5,6,7,9\}\); \(C_4:\{0,1,2,3,6,9\}\) \\
\hline
\end{tabular}
\end{table*}

\begin{table*}[t]
\caption{Deterministic Fashion-MNIST client label-support configurations.}
\label{tab:appendix-fashion-supports}
\centering
\scriptsize
\setlength{\tabcolsep}{4pt}
\renewcommand{\arraystretch}{1.18}
\begin{tabular}{c c c p{0.72\textwidth}}
\hline
\(d\) & \(\mathbf{m}\) & \(f(d)\) & Client label supports \\
\hline
1 & \((5)\) & 0.2000 &
\(C_{0\text{-}4}:\{0,1,2,3,4,5,6,7,8,9\}\) \\
2 & \((3,2)\) & 0.8333 &
\(C_{0\text{-}2}:\{0,1,2,3,4,5\}\); \(C_{3\text{-}4}:\{6,7,8,9\}\) \\
3 & \((2,2,1)\) & 2.0000 &
\(C_{0\text{-}1}:\{0,2,3,5,7,8\}\); \(C_{2\text{-}3}:\{1,4,6,7,8,9\}\); \(C_4:\{0,1,2,3,4,6\}\) \\
4 & \((2,1,1,1)\) & 3.5000 &
\(C_{0\text{-}1}:\{0,1,3,5,6\}\); \(C_2:\{2,6,7,8,9\}\); \(C_3:\{0,4,5,7,9\}\); \(C_4:\{1,2,3,4,8\}\) \\
5 & \((1,1,1,1,1)\) & 5.0000 &
\(C_0:\{0,3,8,9\}\); \(C_1:\{0,3,4,7\}\); \(C_2:\{2,6,7,8\}\); \(C_3:\{1,4,5,6\}\); \(C_4:\{1,2,5,9\}\) \\
\hline
\end{tabular}
\end{table*}

\subsection{Validation Accuracy}

\begin{figure*}[t]
\centering
\begin{minipage}[t]{0.49\textwidth}
\centering
\includegraphics[width=\linewidth]{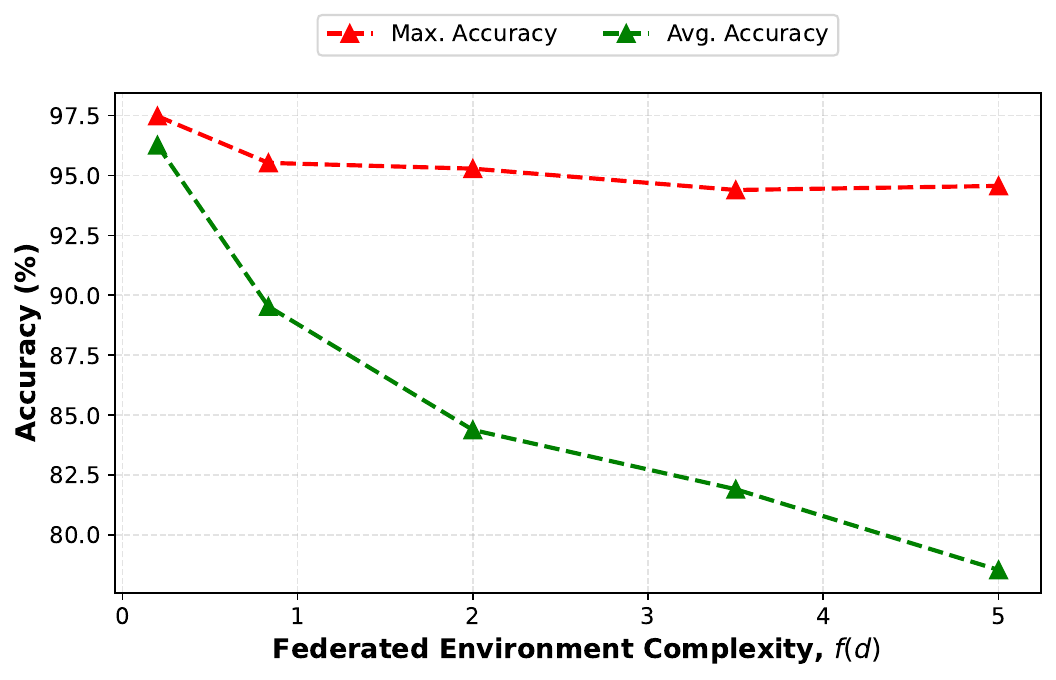}\\[-2pt]
\textbf{(a)} Handwritten-MNIST
\end{minipage}\hfill
\begin{minipage}[t]{0.49\textwidth}
\centering
\includegraphics[width=\linewidth]{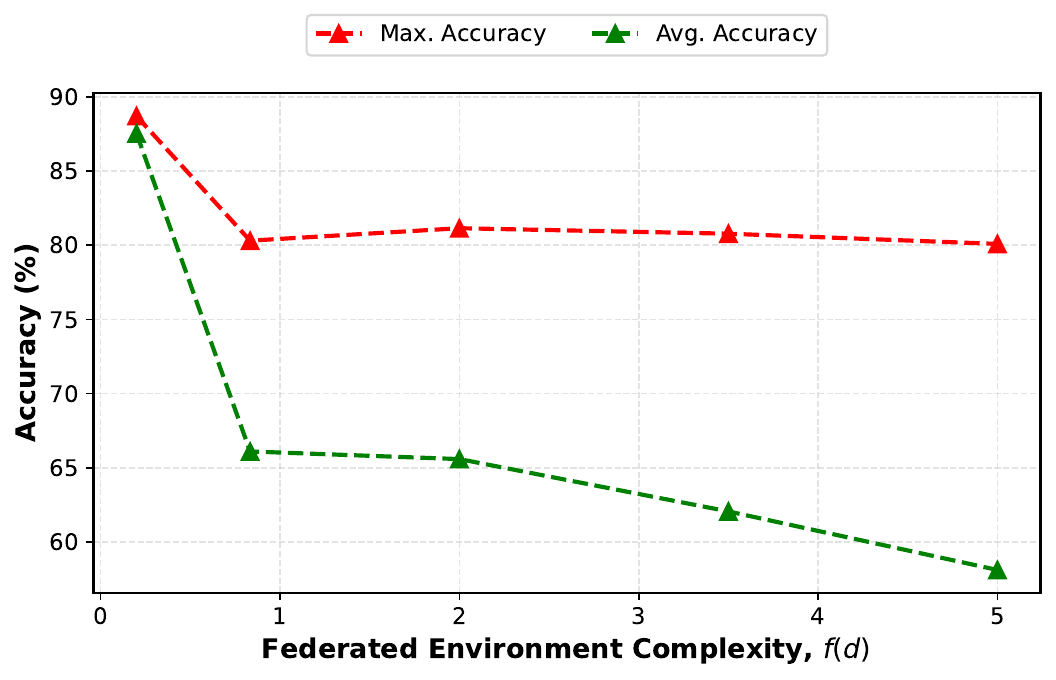}\\[-2pt]
\textbf{(b)} Fashion-MNIST
\end{minipage}
\caption{Federated environment complexity \(f(d)\) versus maximum and average
federated-learning accuracy for the deterministic validation configurations.
The intrinsic-complexity proxy is fixed within each dataset.}
\label{fig:appendix-validation-accuracy}
\end{figure*}

For each dataset and value of \(d\), maximum accuracy is the highest test
accuracy observed across the 100 communication rounds, average accuracy is
the mean test accuracy across those rounds, and final accuracy is the test
accuracy at round 100. Table~\ref{tab:appendix-validation-results} gives the
complete numerical results. Figure~\ref{fig:appendix-validation-accuracy}
shows the maximum- and average-accuracy curves used in the validation.

\begin{table*}[t]
\caption{Accuracy results for the deterministic validation configurations. The combined score is
\(F(d,c)=f_{\mathrm{proxy}}(c)+f(d)\), with \(f_{\mathrm{proxy}}(c)=3.5833\)
for Handwritten-MNIST and \(9.2048\) for Fashion-MNIST.}
\label{tab:appendix-validation-results}
\centering
\small
\setlength{\tabcolsep}{7pt}
\renewcommand{\arraystretch}{1.12}
\begin{tabular}{l c c c r r r}
\hline
Dataset & \(d\) & \(f(d)\) & \(F(d,c)\) & Max. acc. (\%) & Avg. acc. (\%) & Final acc. (\%) \\
\hline
Handwritten-MNIST & 1 & 0.2000 & 3.7833 & 97.47 & 96.2566 & 97.47 \\
Handwritten-MNIST & 2 & 0.8333 & 4.4166 & 95.52 & 89.5351 & 89.72 \\
Handwritten-MNIST & 3 & 2.0000 & 5.5833 & 95.28 & 84.3903 & 92.67 \\
Handwritten-MNIST & 4 & 3.5000 & 7.0833 & 94.39 & 81.9171 & 93.66 \\
Handwritten-MNIST & 5 & 5.0000 & 8.5833 & 94.56 & 78.5540 & 89.63 \\
\hline
Fashion-MNIST & 1 & 0.2000 & 9.4048 & 88.71 & 87.5351 & 88.66 \\
Fashion-MNIST & 2 & 0.8333 & 10.0381 & 80.31 & 66.0957 & 58.55 \\
Fashion-MNIST & 3 & 2.0000 & 11.2048 & 81.15 & 65.5806 & 80.26 \\
Fashion-MNIST & 4 & 3.5000 & 12.7048 & 80.78 & 62.0530 & 65.04 \\
Fashion-MNIST & 5 & 5.0000 & 14.2048 & 80.09 & 58.1045 & 68.01 \\
\hline
\end{tabular}
\end{table*}

\subsection{Combined-Complexity Relationship}

\begin{figure*}[t]
\centering
\includegraphics[width=0.96\textwidth]{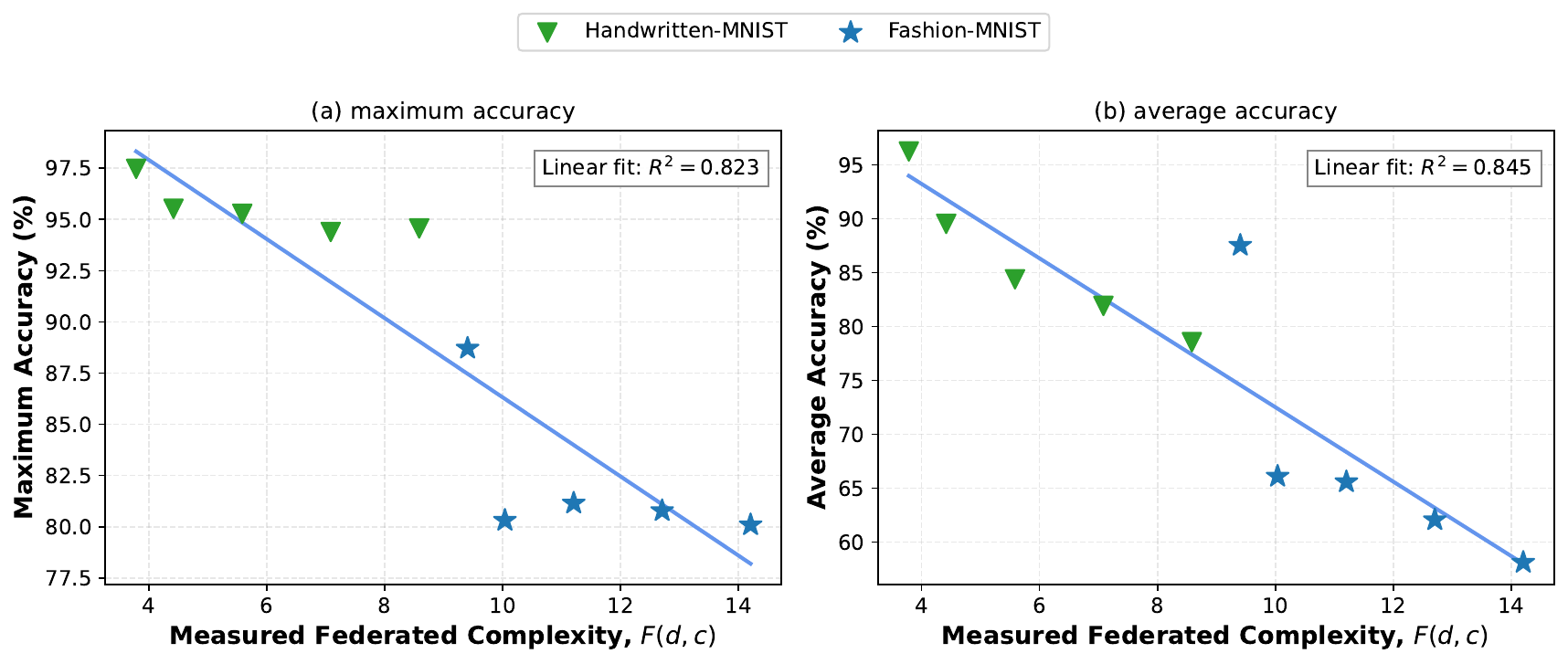}
\caption{Combined federated complexity \(F(d,c)\) versus (a) maximum and
(b) average federated-learning accuracy across the 10 deterministic
Handwritten-MNIST and Fashion-MNIST validation configurations. Lines show
ordinary least-squares fits; the displayed \(R^2\) values equal the squared
Pearson correlations used in the main text.}
\label{fig:appendix-validation-regression}
\end{figure*}

Pooling the five Handwritten-MNIST and five Fashion-MNIST configurations
produces the 10-point validation shown in
Figure~\ref{fig:appendix-validation-regression}. The combined score exhibits a
strong negative relationship with maximum accuracy
\((r=-0.90705,\ R^2=0.82274)\) and average accuracy
\((r=-0.91931,\ R^2=0.84513)\). These values reproduce the rounded
\(R^2\approx0.823\) and \(0.845\) robustness results reported in the main
text. The agreement shows that the combined-complexity relationship persists
under this explicit deterministic support realization, including the mild
non-monotonic variation visible in individual within-dataset accuracy curves.

%% file: reference.bib
@article{Pereyda2020MeasuringDC,
  title={Measuring the complexity of domains used to evaluate ai systems},
  author={Pereyda, Christopher and Holder, Lawrence},
  journal={arXiv preprint arXiv:2010.01985},
  year={2020}
}

@ARTICLE{Scheidegger2021,
  title     = "Efficient image dataset classification difficulty estimation for
               predicting deep-learning accuracy",
  author    = "Scheidegger, Florian and Istrate, Roxana and Mariani, Giovanni
               and Benini, Luca and Bekas, Costas and Malossi, Cristiano",
  journal   = "Vis. Comput.",
  publisher = "Springer Science and Business Media LLC",
  volume    =  37,
  number    =  6,
  pages     = "1593--1610",
  month     =  jun,
  year      =  2021,
  copyright = "https://creativecommons.org/licenses/by/4.0",
  language  = "en"
}

@article{Krusinga2019,
  title={Understanding the (un) interpretability of natural image distributions using generative models},
  author={Krusinga, Ryen and Shah, Sohil and Zwicker, Matthias and Goldstein, Tom and Jacobs, David},
  journal={arXiv preprint arXiv:1901.01499},
  year={2019}
}

@INPROCEEDINGS{Cardoso2021,
  title={Using novelty search to explicitly create diversity in ensembles of classifiers},
  author={Cardoso, Rui P and Hart, Emma and Kurka, David Burth and Pitt, Jeremy V},
  booktitle={Proceedings of the Genetic and Evolutionary Computation Conference},
  pages={849--857},
  year={2021}
}

@ARTICLE{Langley2020,
  title     = "Open-world learning for radically autonomous agents",
  author    = "Langley, Pat",
  journal   = "Proc. Conf. AAAI Artif. Intell.",
  publisher = "Association for the Advancement of Artificial Intelligence
               (AAAI)",
  volume    =  34,
  number    =  09,
  pages     = "13539--13543",
  month     =  apr,
  year      =  2020
}

@article{Batty2014,
  doi = {10.1007/s10109-014-0202-2},
  url = {https://doi.org/10.1007/s10109-014-0202-2},
  year = {2014},
  month = sep,
  publisher = {Springer Science and Business Media {LLC}},
  volume = {16},
  number = {4},
  pages = {363--385},
  author = {Michael Batty and Robin Morphet and Paolo Masucci and Kiril Stanilov},
  title = {Entropy,  complexity,  and spatial information},
  journal = {Journal of Geographical Systems}
}

@article{deng2012mnist,
  title={The mnist database of handwritten digit images for machine learning research},
  author={Deng, Li},
  journal={IEEE Signal Processing Magazine},
  volume={29},
  number={6},
  pages={141--142},
  year={2012},
  publisher={IEEE}
}

@misc{xiao2017fashionmnist,
  author = {Xiao, Han and Rasul, Kashif and Vollgraf, Roland},
  description = {Fashion-MNIST: a Novel Image Dataset for Benchmarking Machine Learning Algorithms},
  title = {Fashion-MNIST: a Novel Image Dataset for Benchmarking Machine Learning
  Algorithms},
  url = {http://arxiv.org/abs/1708.07747},
  year = 2017
}

@ARTICLE{FL2020,
  author={Li, Tian and Sahu, Anit Kumar and Talwalkar, Ameet and Smith, Virginia},
  journal={IEEE Signal Processing Magazine}, 
  title={Federated Learning: Challenges, Methods, and Future Directions}, 
  year={2020},
  volume={37},
  number={3},
  pages={50-60},
  doi={10.1109/MSP.2020.2975749}
  }

@misc{emnist2017,
  doi = {10.48550/ARXIV.1702.05373},
  url = {https://arxiv.org/abs/1702.05373},
  author = {Cohen,  Gregory and Afshar,  Saeed and Tapson,  Jonathan and van Schaik,  André},
  title = {EMNIST: an extension of MNIST to handwritten letters},
  publisher = {arXiv},
  year = {2017},
  copyright = {arXiv.org perpetual,  non-exclusive license}
}

@article{an2020ensemble,
  title={An ensemble of simple convolutional neural network models for MNIST digit recognition},
  author={An, Sanghyeon and Lee, Minjun and Park, Sanglee and Yang, Heerin and So, Jungmin},
  journal={arXiv preprint arXiv:2008.10400},
  year={2020}
}

@inproceedings{tanveer2021fine,
  title={Fine-tuning darts for image classification},
  author={Tanveer, Muhammad Suhaib and Khan, Muhammad Umar Karim and Kyung, Chong-Min},
  booktitle={2020 25th International Conference on Pattern Recognition (ICPR)},
  pages={4789--4796},
  year={2021},
  organization={IEEE}
}

@article{viswanathanwavemix,
     title={WaveMix: A Resource-efficient Neural Network for Image Analysis}, 
      author={Pranav Jeevan and Kavitha Viswanathan and Anandu A S and Amit Sethi},
      year={2024},
      eprint={2205.14375},
      archivePrefix={arXiv},
      primaryClass={cs.CV},
      url={https://arxiv.org/abs/2205.14375}, 
}

@inproceedings{44822,
title	= {Communication-Efficient Learning of Deep Networks from Decentralized Data},
author	= {H. Brendan McMahan and Eider Moore and Daniel Ramage and Seth Hampson and Blaise Aguera y Arcas},
year	= {2017},
URL	= {http://arxiv.org/abs/1602.05629},
booktitle	= {Proceedings of the 20th International Conference on Artificial Intelligence and Statistics (AISTATS)}
}

@article{levina2004maximum,
  title={Maximum likelihood estimation of intrinsic dimension},
  author={Levina, Elizaveta and Bickel, Peter},
  journal={Advances in neural information processing systems},
  volume={17},
  year={2004}
}

@article{shannon1948mathematical,
  title={A mathematical theory of communication},
  author={Shannon, Claude Elwood},
  journal={The Bell system technical journal},
  volume={27},
  number={3},
  pages={379--423},
  year={1948},
  publisher={Nokia Bell Labs}
}

@article{reddi2020adaptive,
  title={Adaptive federated optimization},
  author={Reddi, Sashank and Charles, Zachary and Zaheer, Manzil and Garrett, Zachary and Rush, Keith and Kone{\v{c}}n{\`y}, Jakub and Kumar, Sanjiv and McMahan, H Brendan},
  journal={arXiv preprint arXiv:2003.00295},
  year={2020}
}

@article{bac2021scikit,
  title={Scikit-dimension: a python package for intrinsic dimension estimation},
  author={Bac, Jonathan and Mirkes, Evgeny M and Gorban, Alexander N and Tyukin, Ivan and Zinovyev, Andrei},
  journal={Entropy},
  volume={23},
  number={10},
  pages={1368},
  year={2021},
  publisher={MDPI}
}

@article{scikit-learn,
 title={Scikit-learn: Machine Learning in {P}ython},
 author={Pedregosa, F. and Varoquaux, G. and Gramfort, A. and Michel, V.
         and Thirion, B. and Grisel, O. and Blondel, M. and Prettenhofer, P.
         and Weiss, R. and Dubourg, V. and Vanderplas, J. and Passos, A. and
         Cournapeau, D. and Brucher, M. and Perrot, M. and Duchesnay, E.},
 journal={Journal of Machine Learning Research},
 volume={12},
 pages={2825--2830},
 year={2011}
}

@article{hoyer2004nonnegative,
  author  = {Patrik O. Hoyer},
  title   = {Non-negative Matrix Factorization with Sparseness Constraints},
  journal = {Journal of Machine Learning Research},
  volume  = {5},
  pages   = {1457--1469},
  year    = {2004},
  url     = {https://www.jmlr.org/papers/v5/hoyer04a.html}
}

@inproceedings{paszke2019pytorch,
  author    = {Adam Paszke and Sam Gross and Francisco Massa and
               Adam Lerer and James Bradbury and Gregory Chanan and
               Trevor Killeen and Zeming Lin and Natalia Gimelshein and
               Luca Antiga and others},
  title     = {PyTorch: An Imperative Style, High-Performance
               Deep Learning Library},
  booktitle = {Advances in Neural Information Processing Systems},
  volume    = {32},
  year      = {2019}
}

@article{harris2020array,
  author  = {Charles R. Harris and K. Jarrod Millman and
             St{\'e}fan J. van der Walt and Ralf Gommers and
             Pauli Virtanen and David Cournapeau and others},
  title   = {Array Programming with NumPy},
  journal = {Nature},
  volume  = {585},
  pages   = {357--362},
  year    = {2020},
  doi     = {10.1038/s41586-020-2649-2}
}
